%% file: root.tex
\documentclass[letterpaper, 10 pt, conference]{ieeeconf}  

\IEEEoverridecommandlockouts                              

\usepackage{xcolor}
\usepackage{comment}
\usepackage{graphics} 
\usepackage{epsfig} 
\usepackage{amsmath} 
\usepackage{amssymb}  
\usepackage{mathtools}
\title{\LARGE \bf From Human Walking to Bipedal Robot Locomotion: \\ Reflex Inspired Compensation on Planned and Unplanned Downsteps
}

\author{Joris Verhagen$^{1}$, Xiaobin Xiong$^{2}$, Aaron D. Ames$^{2}$, and Ajay Seth$^{1}$
\thanks{$^{1}$Authors are with the Faculty of Mechanical, Maritime, and Materials Engineering (3ME) and with the faculty of Biomechanical Engineering,
        Delft University of Technology, 2628 CD Delft, The Netherlands
        {\tt\small j.p.m.verhagen@student.tudelft.nl}}%
\thanks{$^{2}$Authors are with the Department of Civil and Mechanical Engineering, California Institute of Technology, Pasadena, CA 91125, USA}
}

\begin{document}

\maketitle
\thispagestyle{empty}
\pagestyle{empty}

\input{0_abstract}
\input{1_introduction}

\input{2_human}

\input{3_RoM}
\input{4_3D}

\input{5_results}
\input{6_conclusions}

\addtolength{\textheight}{-0cm}   

\section{Acknowledgement}
The authors would like to thank Guoping Zhao from the Lauflabor lab at Technische Universit\"at Darmstadt for measurement data and advising on its processing. Xiaobin Xiong would like to thank Helei Duan for his guidance on using the open-sourced simulator of Cassie in Mujoco.
\bibliographystyle{IEEEtran}
\bibliography{ref}


\end{document}

%% file: 0_abstract.tex
\begin{abstract}

Humans are able to negotiate downstep behaviors---both planned and unplanned---with remarkable agility and ease.  The goal of this paper is to systematically study the translation of this human behavior to bipedal walking robots, even if the morphology is inherently different. 
Concretely, we begin with human data wherein planned and unplanned downsteps are taken.  
We analyze this data from the perspective of reduced-order modelling of the human, encoding the center of mass (CoM) kinematics and contact forces, which allows for the translation of these behaviors into the corresponding reduced-order model of a bipedal robot.  We embed the resulting behaviors into the full-order dynamics of a bipedal robot via nonlinear optimization-based controllers.  The end result is the demonstration of planned and unplanned downsteps in simulation on an underactuated walking robot. 



\end{abstract}

%% file: 1_introduction.tex
\section{Introduction}


Bipedal robotics has experienced tremendous progress in the last decades.  Yet, even in fully known environments, the agility and robustness of mechanical bipeds has yet to match their biological counterparts. We argue that this is due to a lack of online (feedback) motion planning, an absence of reflex-like control, and non-advantageous passive dynamics, which biological systems employ to overcome disturbances. 
This can most notably be seen when considering significant unplanned changes in stepping height. For example, \cite{daley2006running} describes the behavior of guinea fowls subjected to an unplanned downstep in a running gait. Similarly, these (un)planned downstep scenarios have been the focus in human running \cite{drama2020postural,ernst2014vertical} and walking \cite{vielemeyer2019ground,van2009hitting}. 
While there has been efforts to achieve similar behaviors in the context of robotic running \cite{green2020planning,vejdani2013bio}, they have yet to be realized on robots with different morphologies.   The goal of this paper, therefore, is to enable the translation of human locomotion behaviors of humans when negotiating both planned and unplanned downsteps to morphologically different bipedal robots.


The nonlinear control and biomechanics communities have traditionally pursued the study of bipedal robot locomotion from different perspectives. 
The control approach is mainly concerned with realizing stable and robust locomotion with formal guarantees---even if the resulting walking does not directly share commonality with human walking. Methods such as offline trajectory optimization with Hybrid Zero Dynamics (HZD) \cite{westervelt2003hybrid,sreenath2011compliant} or closed-form template-model stepping methods \cite{xiong20213d} require a varying degree of model knowledge, but have been successfully utilized to achieve dynamic walking behaviors experimentally on underactuated robots. Roboticists have enabled autonomous bipeds to overcome these downsteps with, for example, switching control \cite{park2012switching}, HZD with a finite state machine \cite{park2012finite}, and Zero Moment Point criteria \cite{takubo2009rough}. However, in contrast to our work, these methods do not consider the advantages that inspiration from biology can offer.
The biomechanist approach typically focuses on actuation- and activation methods, and human morphology. Although formal notions of `biologically-inspired walking' exist \cite{ames2014human}, the intersection of these distinct fields has received less focus than one might expect from the significant similarities between analyzing biological bipedal locomotion and realizing robotic bipedal locomotion. 
Additionally, while there have been approaches to benchmarking human likeness \cite{torricelli2020benchmarking}, this does not address how to achieve human-like behaviors on walking robots. 
In this work, we identify similarity between human and robotic walking via \emph{reduced-order models (RoMs)} and use this to embed human walking behaviors, and specifically downstepping, on walking robots.

\begin{figure}[t]
    \centering
    \framebox{\includegraphics[width= 1\linewidth]{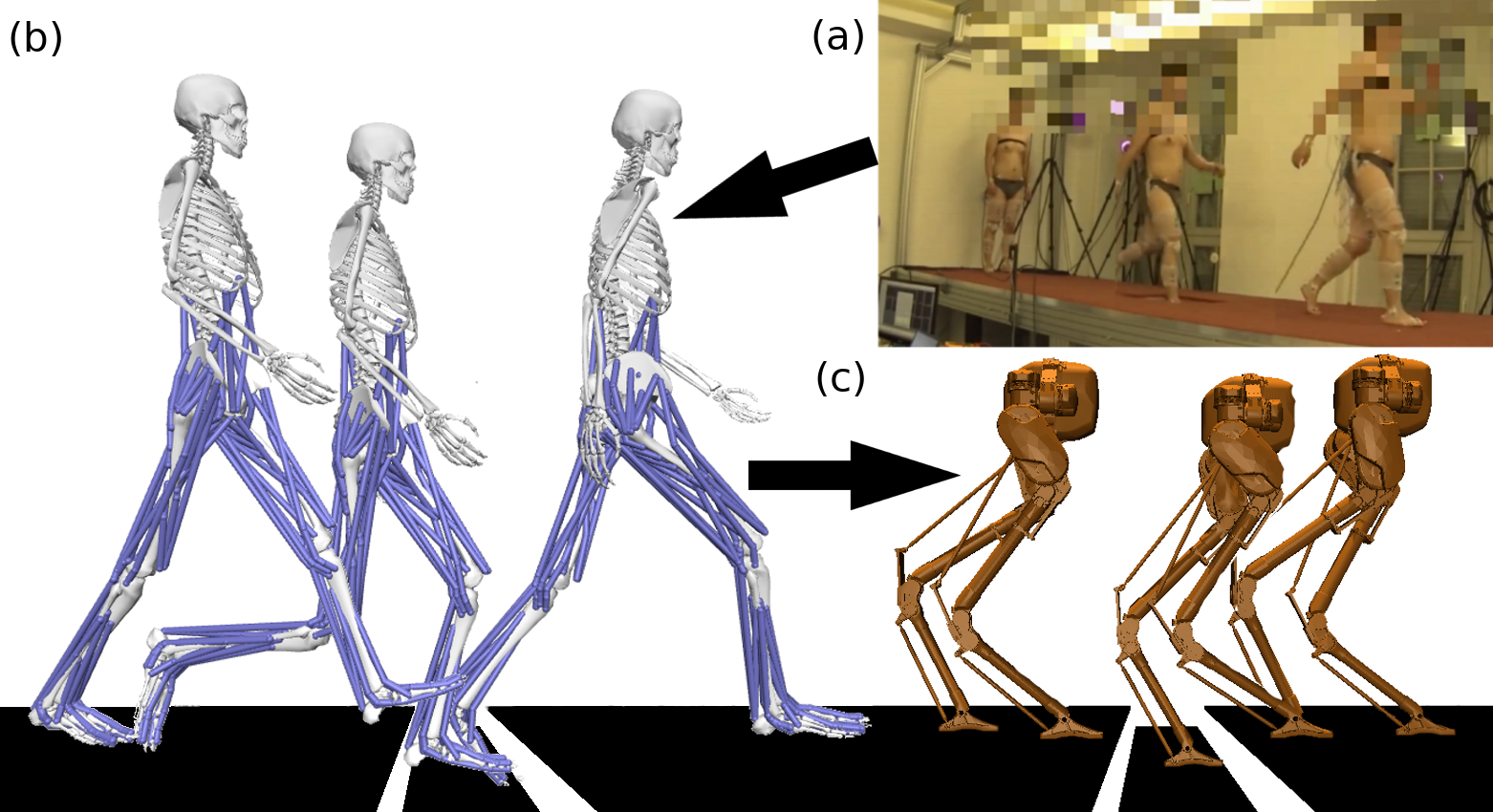}}
    \caption{The human measurement data in OpenSim mapped to a representative skeletal model of the test subject, compared to 3D Cassie subjected to the same downstep height. Changes occur in step-time, step-length, forward walking velocity, desired contact forces, and vertical CoM trajectories.}
    \label{fig:HLIP}
\end{figure}

This paper presents a method for translating downstep behaviors---both planned and unplanned---from humans to walking robots.  Specifically, the 3D bipedal robot Cassie \cite{AG} which is substantially morphologically different from a human as shown in Fig. \ref{fig:HLIP}. Cassie misses an upper body with arms which severely limits the capabilities of changing the angular momentum around the stance leg without changing the stepping behavior.
To achieve this goal, we first use data collected from human walking downstep experiments and abstract this behavior to a reduced-order model (RoM) that captures the essential components of this behavior: the kinematics of the center of mass (CoM) and the ground reaction forces.  We then consider the dynamics associated with this RoM via the Spring-Loaded Inverted Pendulum (SLIP) model with actuation \cite{xiong2018bipedal, xiong2019exo} and generate nominal downstep compensation. We stabilize the vertical state and realize force-embedding with the Backstepping-Barrier Function Quadratic Program (BBF-QP) framework developed in \cite{xiong2021ral} and stabilize the horizontal state via step-size adaptation of the Hybrid Linear Inverted Pendulum (H-LIP) \cite{xiong20213d, xiong2021icra} due to its linear Step-to-Step (S2S) dynamics. For the 3D implementation, a rigid model is assumed where the output dynamics are stabilized towards the desired trajectories using a Task-Space Controller (TSC) with force-embedding as a linear constraint.
The end result of this approach is the ability to generate downstep behaviors in simulation on the 3D model of Cassie.  We, therefore, are able to start from human data of downstepping and, through a principled abstraction of the key elements of locomotion, arrive at robotic downstepping even when the morphologies of the human and robot differ dramatically.

The structure of this paper is as follows.  In Section \ref{sec:human} we describe the human walking on downsteps and corresponding data analysis; specifically related to the CoM and ground reaction forces.  Section \ref{sec:RoM} addresses the dynamic approximation of the human via reduced-order modelling and the realization of the original walking behavior on the RoM via decoupled controllers (BBF-QP with H-LIP stepping). This walking on reduced-order models is embedded into the full-order dynamics of the 3D robot in Section \ref{sec:3Dwalk}.  Finally, the results of the paper are described in Section \ref{sec:results} wherein the 3D walking achieved on Cassie for downstep behaviors---both planned and unplanned---is described.

%% file: 2_human.tex
\begin{figure}[t]
    \centering
    \framebox{\includegraphics[width=0.9\linewidth]{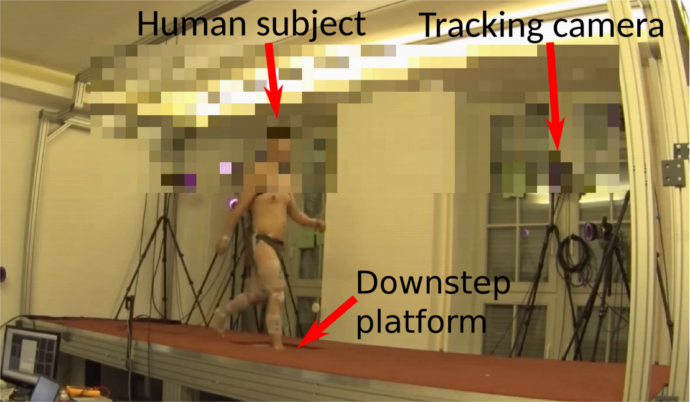}}
    \caption{The experiment setup for data collection of human walking. Force plates are installed beneath the platform and motion capture markers are present on the human subject.}
    \label{fig:setup}
\end{figure}

\section{Human Walking on Downsteps}
\label{sec:human}

\subsection{Data Collection}
To understand (un)planned downsteps in humans, we analyzed experiments conducted at the Lauflabor lab of the Technische Universit\"at Darmstadt. Human subjects were instructed to walk on a platform, 2 meters wide and 6 meters long, shown in Fig. \ref{fig:setup}. A variable height walking platform is located at the center of the walking platform. Three force plates are present, before, on, and after the variable height platform, that record the ground reaction forces at 1 kHz. Full body movement is recorded by a motion capture system consisting of 26 markers and 16 cameras at 240 Hz. Eight trials are conducted for each downstep scenario with heights equal to 0.0 cm, -2.5 cm, -5.0 cm, -7.5 cm, and -10.0 cm for both planned and unplanned situations\footnote{All trials allow full vision of the walking platform by the participant (no partial blocking is needed). Unplanned trials are performed by suddenly lowering the walking platform when the swing foot approaches the ground.}. A total of nine experimental conditions are considered: four known downstep heights, four unknown downstep heights, and nominal flat-ground walking.

Inverse Kinematics (IK) optimization is performed on a representative musculoskeletal model in OpenSim \cite{delp2007opensim,seth2018opensim} to obtain the joint angles and Center of Mass (CoM) positions and velocities that best fit the marker trajectories. 
Differences in behavior in the coronal plane are deemed to be small and insufficiently affected by the downsteps. Subsequently, we limit our focus to the results in the sagittal plane.


\subsection{Kinematics and Kinetics Analysis}
\begin{figure}[t]
    \centering
    \framebox{\includegraphics[width=0.9\linewidth]{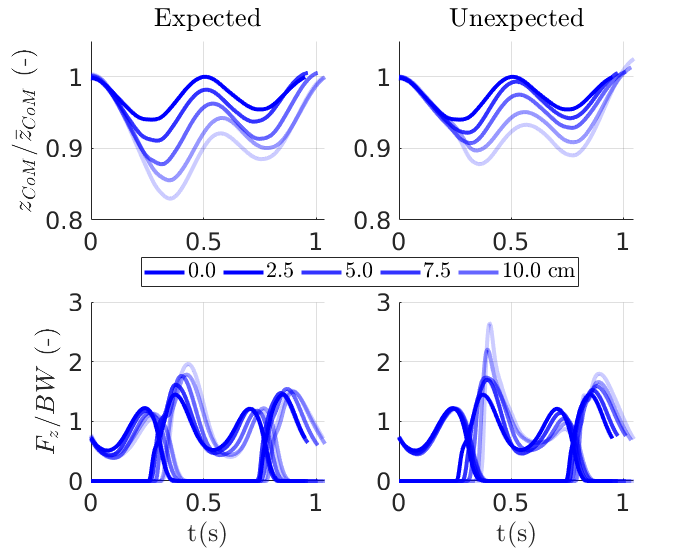}}
    \caption{Average vertical CoM position normalized by the CoM height at VLO and vertical GRF normalized by the body weight for flat-ground, planned, and unplanned downsteps. Lighter shades indicate an increase in the downstep height from flat-ground (0 centimeter) to 10 centimeter. The GRFs start at the VLO before the downstep, switch to the swing leg experiencing the downstep, and ends at the original stance leg when the downstep has been overcome.}
    \label{fig:com_trials_00_10_10}
\end{figure}
A powerful abstraction for a full-body analysis of the human gait is on the CoM behavior. Although the human subjects have an upper body with arms (which are not present on Cassie), the contributions towards its angular momentum around the stance leg can nevertheless be captured by the point-mass dynamics. Fig \ref{fig:com_trials_00_10_10} shows a polynomial fit to the mean vertical CoM position for the nominal and downstep trials and the ground reaction force (GRF) for the leg that experiences the downstep. The analysis is performed from the Vertical Leg Orientation (VLO) before the downstep and ends at the VLO after the downstep\footnote{Two complete steps are considered. If we consider a left stance leg during downstep detection, the analysis is from the moment of the CoM passing the left foot on the raised pre-downstep platform until the moment of the CoM passing the left foot on the raised post-downstep platform.}.

\begin{figure}[t]
    \centering
    \framebox{\includegraphics[width=0.9\linewidth]{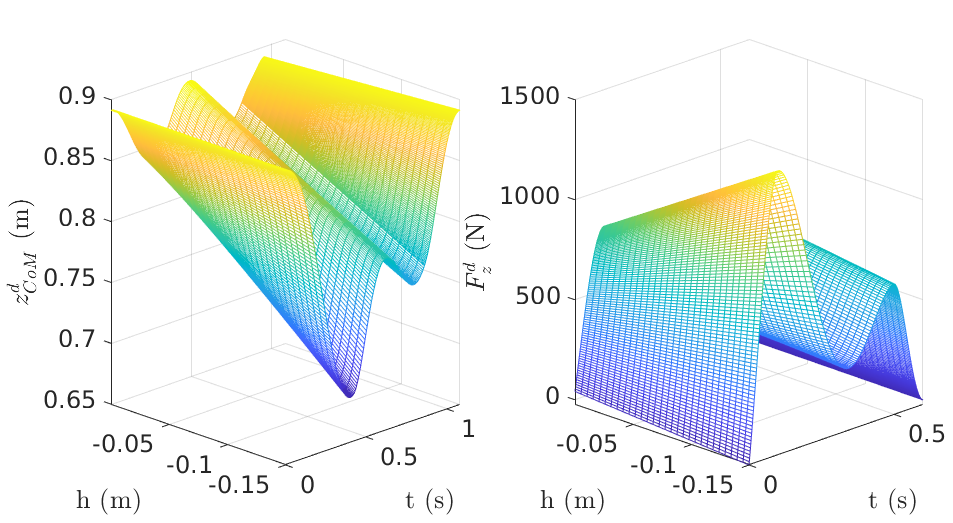}}
    \caption{The desired vertical CoM trajectory $z^d_{CoM}$ and the GRF of the leg that experiences the downstep parameterized by time ($t$) and the downstep height $h$ for planned downsteps. Similar regressions are applied for the walking on unplanned downsteps and other legs for both the Single Support Phase (SSP) and Double Support Phase (DSP)}
    \label{fig:surface_plots}
\end{figure}

The variance between the normalized results of the three participants is low. From Fig \ref{fig:com_trials_00_10_10}, we observe that the vertical CoM position is significantly lowered before the swing foot impacts the downstep platform. This is especially the case for planned downsteps where this already occurs from VLO before the downstep onwards. For these planned downsteps, the CoM height is lowered more significantly while the impact force on the downstep platform is reduced during. For unplanned downsteps, the change in vertical CoM height during downstep is predominantly caused by the passive pendulum properties of the stance leg, and the peak of the GRF is significantly higher compared to the planned downsteps. From the measurement data, we create continuously differentiable ($C^1$) surfaces for the vertical CoM position and the desired GRFs shown in Fig. \ref{fig:surface_plots}.

\begin{figure}[t]
    \centering
    \framebox{\includegraphics[width=0.9\linewidth]{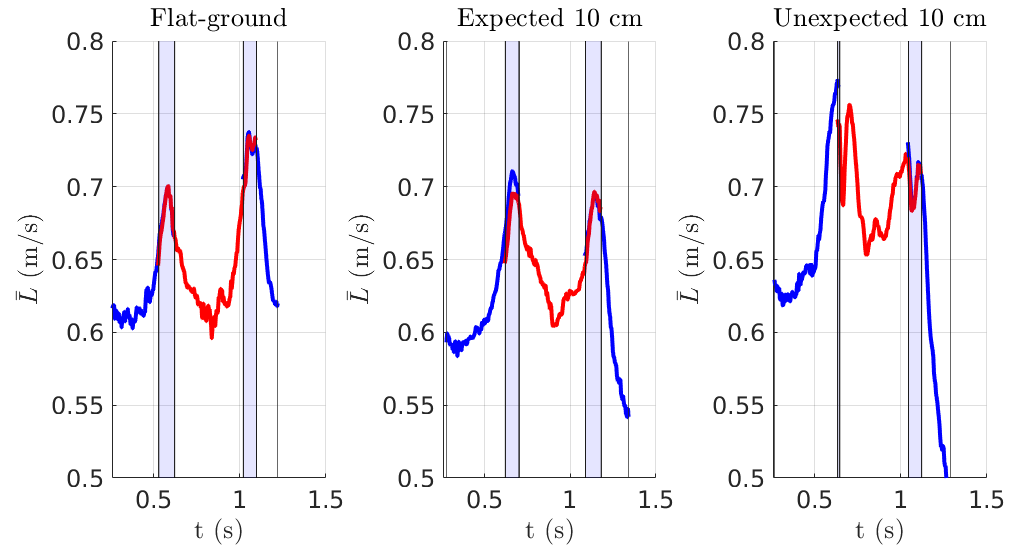}}
    \caption{Averaged trajectories of the angular momentum around the contact point for walking on flat-ground, planned, and unplanned downsteps (with 10cm depth). The leg experiencing the downstep is indicated in red, the other leg in blue. Blue shaded regions indicate the DSP.}
    \label{fig:angular_momentum_00_10_10}
\end{figure}

With the subjects being instructed to `continue' their gait, an important metric with regards to their stability is the angular momentum around the stance leg. The stance-foot angular momentum is shown in Fig \ref{fig:angular_momentum_00_10_10}. These results indicate that the angular momentum is much more contained towards the flat-ground walking condition for planned trials. This is caused both by the reduced vertical CoM position and the smaller change in horizontal CoM velocity. The latter is caused by an increase in step-size as it was noticed that changes to the nominal step-lengths in unplanned downstep trials were governed by the passive dynamics of the swing leg.

\subsection{Human Walking Model Reduction}
\label{ssec:aSLIP_optimization}

The contributions of muscle activation (either deliberate or reflexive) and changes to the posture alter the dynamic behavior of the human when subjected to planned and unplanned downsteps. The subsequent analysis would be high dimensional. Additionally, our results from the measurement data are noisy and do not explicitly contain the acceleration of the CoM. Lastly, these results are only to an assumed extend representative of a point-mass model. In order to obtain a tractable analysis of the dynamics, we abstract the human towards the actuated Spring Loaded Inverted Pendulum (aSLIP) model from \cite{xiong2018bipedal, xiong2021ral} which contains damping and actuation on the rest-length of the spring, compared to the canonical SLIP model. 

We construct a non-convex optimization problem of fitting the behavior of the human to the aSLIP model. With the introduction of the actuation, we will jointly optimize a leg-length dependent stiffness and a constant damping for nominal walking (and use this in the downstep walking scenarios). We optimize  the acceleration of the rest-length of the springs when in contact with the ground (which is our actuation signal $\ddot{L}$) for all scenarios. The changes to the representative stiffness of the human leg via deliberate and reflexive action and changes in posture are therefore encaptured by the change in the physical length and the rest-length of the spring. The optimization is formulated as
\begin{align}
\label{eq:direct_collocation}
    \min & \sum_{i=1}^5 (||z^a_{\text{CoM}} - z^d_{\text{CoM}}||^2 + w(||\ddot{L}_1||^2 + ||\ddot{L}_2||^2)) \\
     \text{s.t.}  &f_{\text{aSLIP}} + g_{\text{aSLIP}}\ddot{L} = 0 \tag{RoM dynamics} \\
          &\mathbf{x}_i(t_f) = \mathbf{x}_{i+1}(t_0), \forall k\in\{0,1,2,3\} \tag{continuity}\\
          &\mathbf{x}_0(t_0) = \mathbf{x}_5(t_f) \tag{VLO}\\
          &F_{z,k}(t) \geq 0,  \forall k,  \forall t \nonumber \\
          &F_{z,sw}(t_0) = 0, \forall k \nonumber \\
          &F_{z,st}(t_f) = 0, \forall k, \nonumber
 \end{align}
where $i$ indicates the phase (SSP or DSP) of the walking gait, $w \in \mathbb {R}$ is a scaling parameter on the cost, $z^a_{\text{CoM}}$ and $z^d_{\text{CoM}}$ are the actual and desired vertical position of the CoM, $\ddot{L}_j$ is the acceleration on the rest-length of leg $j$, $f_{\text{aSLIP}}$ and $g_{\text{aSLIP}}$ represent the equations of motion of the aSLIP model in either SSP or DSP, $\mathbf{x}_i$ indicates the full state of the system at phase $i$, $F_{z,k}$ is the vertical GRF at phase $k$, and  $F_{z,st}$ and $F_{z,sw}$ are the vertical GRFs of the current stance and new stance leg respectively. As we consider the analysis from pre-downstep VLO to post-downstep VLO, we jointly optimize 5 $\text{phases}^2$. Consistency in the horizontal velocities is enforced with soft bounds on the step duration as to not over-constrain the dynamics of the RoM. From the optimization for the nominal walking condition, we also obtain the leg length-dependent stiffness, for which we assume a second degree polynomial as shown in Fig. \ref{fig:human_aSLIP_stiffness}, and the damping, which is assumed constant. Higher degrees of parameterizations of the stiffness and damping were evaluated in the same optimization framework, which are not showing significant improvement on lowering the cost. 


%% file: 3_RoM.tex
\section{Walking Regeneration on a Reduced-Order Model of Humans}
\label{sec:RoM}


Given the kinematics and kinetics data of human walking, we first want to re-generate the motion via feedback control on the optimized aSLIP model (with the stiffness and damping from the optimization) that represents human walking dynamics for all scenarios. We apply the Backstepping-Barrier Function (BBF) controller with the step-to-step (S2S) dynamics approximation approach developed in \cite{xiong2021ral}. The backstepping component in the BBF based quadratic program (BBF-QP) allows the tracking of the vertical state of the point mass, which is underactuated due to the spring in the leg; the control Barrier function in the BBF-QP allows the GRF to stay in a range around the desired GRF profile of human walking. The S2S dynamics approximation provides stepping stabilization that addresses the point-foot underactuation of walking.

\subsection{Vertical CoM Tracking}
\begin{figure}[t]
    \centering
    \framebox{\includegraphics[width=0.95\linewidth]{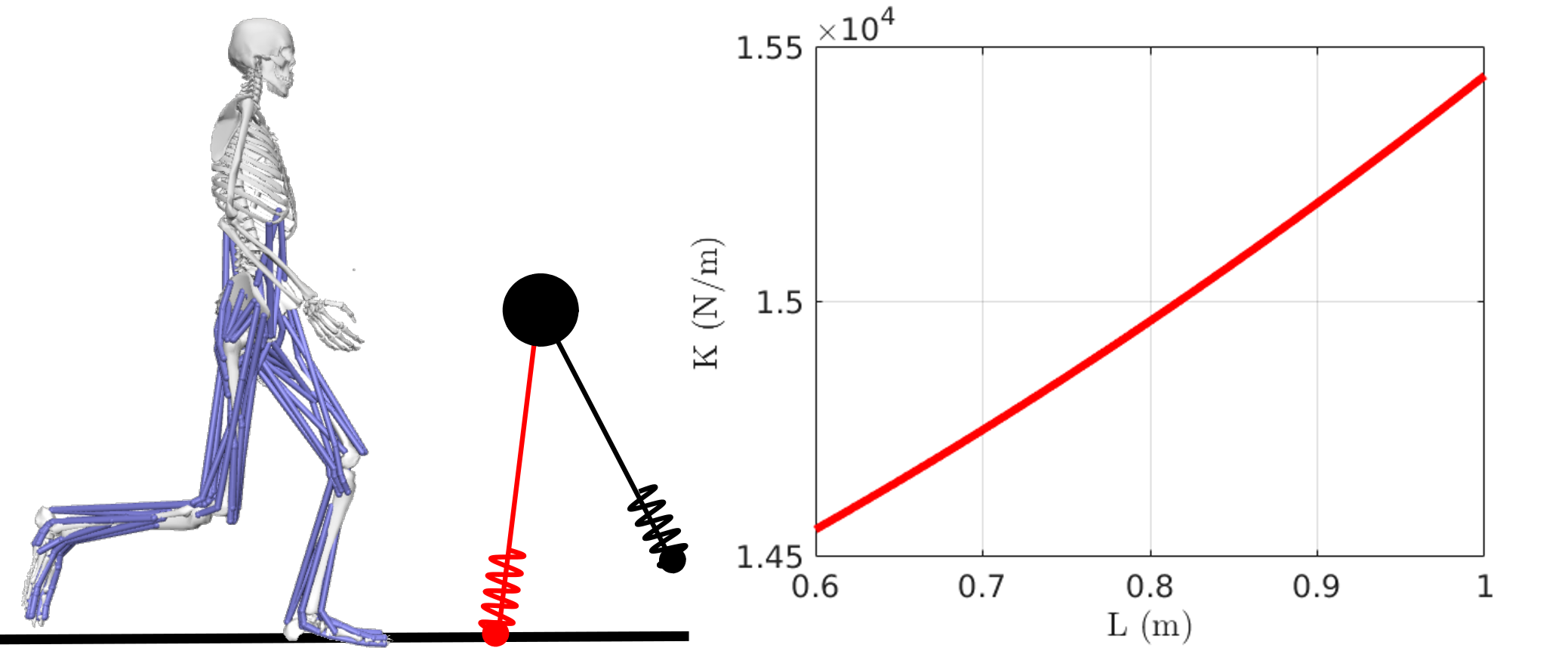}}
    \caption{Abstraction of the human kinematics and kinetics towards the reduced order aSLIP model of walking. We optimize a quadratic stiffness $K(L)$ and a constant damping $D$ which minimizes Eq. \eqref{eq:direct_collocation} for the flat-ground walking condition which is parameterized by the variable rest length $L$ of the spring.
    \label{fig:human_aSLIP_stiffness}}
\end{figure}
For the vertical state, we define the objective as driving the vertical CoM position to follow a desired trajectory during nominal walking and during (un)planned downsteps. These trajectories are obtained by traversing the $C^1$ surface from Fig. \ref{fig:surface_plots}. The output is defined as: 
\begin{equation}
    \eta = \begin{bmatrix} z^a_{\text{CoM}} - z^d_{\text{CoM}}(t) \\ \dot{z}_{\text{CoM}} - \dot{z}_{\text{CoM}}^d(t) \end{bmatrix},
\end{equation}
which leads to the definition of the output dynamics as
\begin{equation}
\label{eq:system_z}
    \dot{\eta} = \begin{bmatrix} \dot{z}^a_{\text{CoM}} - \dot{z}_{\text{CoM}}^d(t) \\ -g - \ddot{z}_{\text{CoM}}^d(t) \end{bmatrix} + \begin{bmatrix} 0 \\ \frac{1}{m} \end{bmatrix} F_{z}^{P} = f_{\eta} + g_{\eta}F_{z}^{P},
\end{equation}
where $F_{z}^{P}$ is the net vertical force on the CoM for each domain $P$ (SSP or DSP). The GRF is related from the spring forces in the leg; e.g., during the SSP, the vertical component of the GRF is:
\begin{equation}
\label{eq:F_z}
    F_z^{\text{SSP}} = (K(L)s + D\dot{s})\cos({\theta_{st}}), 
\end{equation}
where $K(L)$ is the rest-length dependent leg stiffness, $D$ is the damping constant, $\theta_{st}$ is the stance leg angle, and $s = L - L_0$ is the spring deformation. Taking the derivative of the vertical GRF w.r.t. time results in the affine control system for which the state is the input to the system Eq. \eqref{eq:system_z}:
\begin{align}
    \dot{\eta} &= f_{\eta} + g_{\eta}F_{z}^{P} \label{eq:strict_feedback_1} \\ 
    \dot{F}_z^{P} &= f_z + g_z \tau_z, \label{eq:strict_feedback_2}
\end{align}
where $\tau = \ddot{L}$ is the acceleration of the rest-length of the aSLIP and $P$ indicates the phase (SSP or DSP) as in Eq. \eqref{eq:system_z}. As such, Eq. \eqref{eq:strict_feedback_2}, when specifically considering SSP, is the first time derivative of Eq. \eqref{eq:F_z}. As this system is in \textit{strict-feedback form}, we can apply a control Lyapunov function version of the canonical backstepping approach to stabilize the dynamics of both systems with the augmented Lyapunov equation $V(\eta,F_z)= \eta^TP\eta + \frac{1}{2}(F_z - \bar{F}_z)$. More details can be seen in \cite{xiong2021ral}. 

We also want to enforce the desired GRF from the human walking in the controller. The time derivative of $F_z$ is affine in the control input $\ddot{L}$ which allows contact force embedding with Control Barrier Functions (CBF) based on the constraint
\begin{equation}
\label{eq:desired_GRF_RoM}
    (1-c)F_{z}^d + \Delta_F \leq F_{z}^a \leq (1+c)F_{z}^d - \Delta_F,
\end{equation}
where $F_{z}^a$ is the actual GRF as the state in Eq. \eqref{eq:strict_feedback_2}, $F_{z}^d$ is the desired GRF obtained from the $C^1$ surface in Fig. \ref{fig:surface_plots}, $c \in (0,1)$ is a relaxation parameter, and $\Delta_F$ is an additional bound such that the permissible set at the boundary of DSP is nontrivial \cite{xiong2021ral}. The CBF can be included in both the SSP (for one leg) and the DSP (for two legs) as linear constraints in the Control Lyapunov Function Quadratic Program (CLF-QP). In \textit{SSP}, we define a single CBF which ensures the robot's GRF remains in a relaxed tube. In \textit{DSP} the former stance foot has a GRF that goes to zero while the former swing foot has a GRF that goes from zero to the initial GRF of the following SSP. During the downstep, we use the GRF trajectory from Fig. \ref{fig:surface_plots} as the desired $F_{z}^d$.

\subsection{Horizontal stabilization}
\label{ssec:horizontal_stabilization}
The horizontal state is stabilized using the S2S dynamics approximation via the Hybrid Linear Inverted Pendulum (H-LIP). Using a constant height assumption on the vertical CoM during SSP and DSP (which is relaxed due to the tracking of the human vertical CoM behavior), the S2S dynamics of the system can be described in closed-form in accordance to \cite{xiong20213d}. In SSP, the horizontal dynamics of the H-LIP model are described by $\ddot{p} = \lambda^2p$, where $p$ is the horizontal position of the CoM w.r.t. the stance foot and $\lambda = \sqrt{g/z_0}$ with $g$ being the gravity constant and $z_0$ being the nominal walking height. In DSP, we assume a constant horizontal CoM velocity. The S2S dynamics (from the end of SSP of step $k$ to the end of SSP of step $k+1$) of the H-LIP are step-size and step-time dependent according to
\begin{equation}
    \begin{multlined}
    \label{eq:HLIP_propagation}
    \mathbf{x}_{\text{SSP}k+1}^- = e^{A_{\text{SSP}}T_{\text{SSP}}}\begin{bmatrix}1 & T_{\text{DSP}} \\ 0 & 1\end{bmatrix}\mathbf{x}_{\text{SSP}k}^- + \\ e^{A_{\text{SSP}}T_{\text{SSP}}}\begin{bmatrix}-1\\0\end{bmatrix}u_k,
    \end{multlined}
\end{equation}
where $\mathbf{x}_{\text{SSP}k}^-$ is the post-impact horizontal CoM state $\mathbf{x}_{\text{SSP}}^- = \begin{bmatrix} p_{\text{SSP}}^- & \dot{p}_{\text{SSP}}^- \end{bmatrix}^T$ at step $k$, $u_{k}$ is the step-size, $T_{\text{SSP}}$ and $T_{\text{DSP}}$ are the duration of the SSP and DSP respectively, and $A_{\text{SSP}}$ originates from the state-space representation of the SSP dynamics of the H-LIP
\begin{equation}
    \frac{d}{dt}\begin{bmatrix}p\\\dot{p}\end{bmatrix} = \begin{bmatrix} 0 & 1 \\ \lambda^2 & 0 \end{bmatrix} \begin{bmatrix} p \\ \dot{p} \end{bmatrix} := A_{\text{SSP}} \begin{bmatrix} p \\ \dot{p} \end{bmatrix} .
\end{equation}
As mentioned previously, in reality we have a non-constant vertical CoM position from the aSLIP nominal- and compensation gait. The contribution of this deviation instead is regarded as a model difference between the H-LIP and the system (human or robot) according to
\begin{equation}
    \mathbf{x}_{k+1} = A \mathbf{x}_{k} + B u_{k} + w.
\end{equation}
The stepsize for flat-ground walking is determined by 
\begin{equation}
    u_k^d = u_k^{\text{H-LIP}} + K_{db}(\mathbf{x}^{\text{aSLIP}} - \mathbf{x}^{\text{H-LIP}}),
\end{equation}
where $u_k^{\text{H-LIP}}$ is the nominal step-size of the H-LIP, $K_{db}$ is the deadbeat gain (i.e. $(A+BK_{db})^2=0$), and $x^{\text{aSLIP}}$ is the horizontal state of the aSLIP walker. More details of the H-LIP stepping can be found in \cite{xiong20213d,xiong2021ral}. For the downstep scenario's, the H-LIP is taking the \textit{slope} of the walking surface into account. For the planned downstep, the slope is altered at the VLO before the downstep based on the previous step-size and the known downstep height. For the unplanned downstep, the slope is altered continuously based on the current step-size and the penetration of the swing foot.



%% file: 4_3D.tex
\begin{figure*}[t]
    \centering
    \framebox{\includegraphics[width=0.98\textwidth]{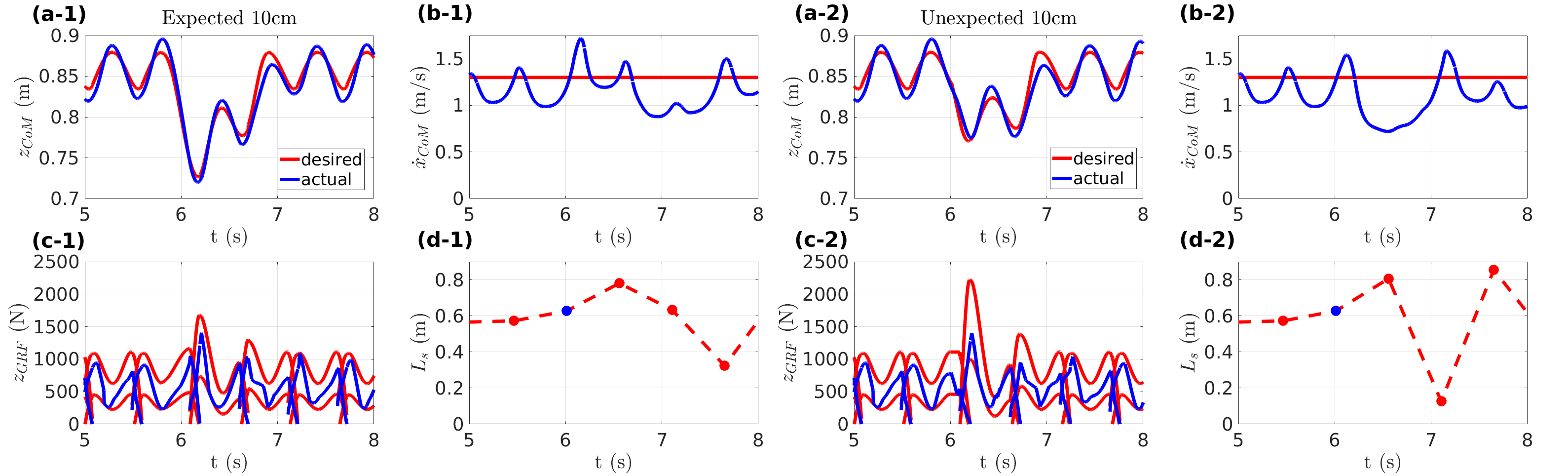}}
    \caption{Simulation results of the aSLIP walking over the planned (1) and unplanned (2) downstep with 10cm depth: (a) the vertical CoM trajectories of the desired ones and the actual ones, (b) the horizontal, forward, velocity of the mass, (c) the GRF profile with its bounds, and (d) the step-lengths}
    \label{fig:human_BBF}
\end{figure*}

\section{3D Robotic Walking Realization}
\label{sec:3Dwalk}
We now present our human inspired walking synthesis on a 3D bipedal robot. We use the robot Cassie as an example, which is a bipedal walking system with significant morphological differences compared to the human. 

\subsection{Human Inspired Trajectory Synthesis}
Before we translate the observed motion and dynamics of the human to a bipedal robot, we first emphasize several differences between the two systems. A robot may have a different distribution of mass and it may not have an upper body or arms. The robot Cassie has a lower CoM. The abstraction towards the CoM assumes the whole-body behavior during walking is primarily captured by the CoM dynamics. A robot may also have different leg compliance compared to the human test subject. The feet of the robot may have underactuated point-feet or only allows flat-footed walking (which we consider is the case for Cassie). This limits the realizable walking behaviors such as the foot rolling that is present in human locomotion. In this work, we do not consider the compliance in the robot or complex foot rolling behaviors on the robot. 

Based on the RoM characterization of the human walking, we want to transfer the CoM trajectory and the GRF profile from the human to the robot Cassie. Firstly, the nominal leg length of Cassie (as defined by the distance between the contact pivot and the CoM, rather than the hip) is a decision variable that determines the scaling of the other gait parameters. For a chosen averaged leg length over a step, we assume the SSP duration of Cassie is related to that of the human by the passive pendulum properties of the swing phase in nominal walking
\begin{equation}
\label{eq:step_time_scaling}
    T_{S,C} = \frac{\sqrt{\bar{L}_{C}/g}}{\sqrt{\bar{L}_{H}/g}}T_{S,H},
\end{equation}
where $T_{S,C}$ and $T_{S,H}$ are the walking period and $\bar{L}_C$ and $\bar{L}_H$ are the averaged leg length of Cassie and the human respectively.
For the flat-footed walking, we remove the horizontal displacement of the CoM caused by the foot-roll phases due to the flat-footed walking on Cassie according to
\begin{equation}
\label{eq:xcom_scaling}
    x_{\text{CoM},C} = \frac{1 - \bar{x}_{\text{roll}}}{L_{s,H}}x_{\text{CoM},H},
\end{equation}
where $x_{\text{CoM},C}$ is the scaled horizontal displacement of the CoM for Cassie, $\bar{x}_{roll}$ is horizontal displacement of the CoM during the heel and toe roll phases normalized by the total step length, $L_{s,H}$ is the leg length of the human, and $x_{\text{CoM},H}$ is the horizontal displacement of the CoM of the human. This foot-roll phase takes a certain amount of time which is removed from the DSP duration of the human and scaled with the fraction from Eq. \eqref{eq:step_time_scaling}. The averaged and overall leg length of Cassie appearing in Eq. \eqref{eq:step_time_scaling} and Eq. \eqref{eq:xcom_scaling} need to be obtained. We do this via the fractional change in virtual leg length of the human and removing the contribution of the roll phases to the stance leg angle in accordance to
\begin{equation}
    L_{s,C} = \frac{L_{s,H} - \bar{L}_H}{\bar{L}_H}\bar{L}_C,
\end{equation}
where, as mentioned, $\bar{L}_C$ is the decision variable of the overall walking height of Cassie.
This allows us to redefine the nominal walking velocity for Cassie as 
\begin{equation}
    \dot{\bar{x}}_{\text{CoM},C} = \frac{x_{\text{CoM},C}(t_f) - x_{\text{CoM},C}(t_0)}{T_{S,C}},
\end{equation}
where $t_0$ and $t_f$ are the start and end-time of a step. The desired ground reaction forces are scaled with the mass fraction according to
\begin{equation}
\label{eq:grf_scaling}
    F_{z,C} = \frac{m_C}{m_H} F_{z,H},
\end{equation}
where $F_{z,C}$ and $F_{z,H}$ are the vertical GRFs and $m_C$ and $m_H$ are the total masses of Cassie and the human respectively.
The resulting outputs can be embedded onto the aSLIP representation obtained in \cite{xiong2018bipedal} which are not shown due to space constraints. For these scaled kinematics and kinetics, similar multi-parameterized surfaces as shown in Fig. \ref{fig:surface_plots} are available for Cassie.

As we want to control the full-order dynamics of Cassie, we are not only concerned with vertical CoM tracking with force-embedding and horizontal stabilization, we also need to control the additional degrees of freedom of Cassie related to the swing foot and the coronal plane. Subsequently, the outputs of the walking are defined as 
\begin{equation}
	y^d =  \begin{bmatrix} \begin{bmatrix}\alpha_{\text{pelvis}} & \beta_{\text{pelvis}} & \gamma_{st} & \gamma_{sw}\end{bmatrix}^T \\ 
						z_{\text{CoM}}^d(t,z_{sw},n_{ds}) \\ 
						x_{sw}(t,\theta) \\ 
						y_{sw}(t,\theta) \\ 
						z_{sw}(t,n_{ds}) \\ 
						\alpha_{sw} \end{bmatrix},
\end{equation}
where $\alpha_{sw}$ and $\gamma_{sw}$ indicate the pitch and yaw of the swing leg which are represented by Bézier splines guiding the trajectory to 0 angle, $n_{ds}$ indicates the downstep step and $\theta$ indicates the slope of the downstep. 
$\theta$ is defined as follows for the different scenarios: for the \textbf{planned} scenario, the downstep height is known in advance and the downward slope is determined by the downstep height and the previous step-size. for the \textbf{unplanned} scenario, the downward slope is determined by the foot penetration of the supposed flat ground and the previous step-size. It is therefore continuously updated when the swing foot travels through the unknown downstep height. For the upwards slope, $\theta$ is known for both planned and unplanned scenarios as the downstep height can be estimated by the robot's joint positions.

The swing foot positions in the horizontal plane originate from a decoupling of a Period-1 H-LIP for the sagittal plane and a Period-2 H-LIP for the coronal plane as shown in \cite{xiong20213d}. The vertical position of the swing foot tracks a pre-defined Bézier spline with unique formulations for the two steps after the initial downstep based on the, now known, downstep height. The desired GRFs are parameterized similarly to the vertical CoM trajectories $F^d(t,z_{sw},n_{ds})$.

\subsection{Contact Force Embedded Task Space Control}

\begin{figure*}[t]
    \centering
    \framebox{\includegraphics[width=0.98\textwidth]{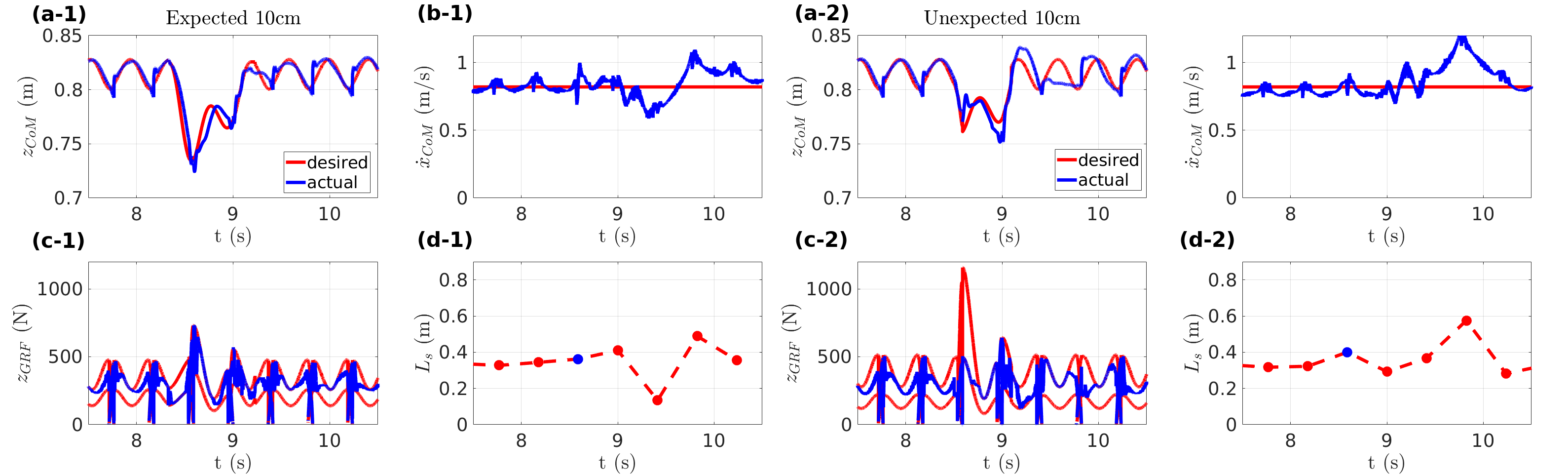}}
    \caption{Simulation results of Cassie walking over the planned and unplanned downsteps with 10 centimeter depth. The individual subfigures are explained in Fig. \ref{fig:human_BBF}. }
    \label{fig:cassie_3D}
\end{figure*}



To realize the proposed trajectory synthesis on the bipedal robot, we apply the task space controller (TSC) for output tracking. The force embedding can be realized via a linear constraint on the holonomic forces, which are optimization variables in the TSC. This is in contrast to the approach with CBFs for the RoM walking realization in Eq. \ref{eq:desired_GRF_RoM}. We directly specify a linear constraint on the vertical GRF to realize the force-embedding
\begin{equation}
	(1-c)F^d + \Delta_F \leq SF_h \leq (1+c)F^d - \Delta_F,
\end{equation}
where $F^d = F^d(t,z_{sw},n_{ds})$ and $S$ is a selection matrix to select the vertical component of the GRF from $F_h$ (the vector of all the holonomic forces).


The final quadratic program with the equation of motion (EOM) constraint, holonomic constraints, contact force constraints is formulated as:
\begin{align}
  \quad  \min_{u,F_h,\ddot{q}} & \ \ ||\ddot{y}^a - \ddot{y}^d - \ddot{y}^t||^2 \\
 \quad   \text{s.t.} \hspace{4mm} & D\ddot{q} + C = J_h^T F_h + B \tau \nonumber  \tag{EOM} \\
                          & J_h \ddot{q} + \dot{J}_h \dot{q} = 0 \nonumber \tag{holonomic}   \\
                                              &\tau_{\text{min}} \leq \tau \leq \tau_{\text{max}} \nonumber \tag{torque limit} \\
                                              &A_\text{GRF}F_\text{GRF} \leq 0\nonumber  \tag{friction cone}\\
                                                 & (1-c)F^d + \Delta_F \leq S F_h \leq (1+c)F^d + \Delta_F  \nonumber  
\end{align}
where $q$ is the configuration, $D$ is the mass matrix, $C$ is the Coriolis and gravitation term, $J_h$ is the Jacobian of the holonomic constraints, $B$ is the actuation matrix, $\tau$ is the input torque, and $A_\text{GRF}$ is a constant matrix that specifies the friction cone constraints. $y^a$ and $y^d$ are the actual and desired outputs respectively, and $\ddot{y}^t = - K_p (y^a - y^d) - K_d (\dot{y}^a - \dot{y}^d)$ with $K_p, K_d$ being the feedback proportional and derivative gain matrices. 

The QP is solved using OSQP \cite{stellato2020osqp} at 2 KHz in the Mujoco physics simulator \cite{todorov2012mujoco,cassiemujocosim}. During SSP and DSP, each constraint in the QP considers one or two feet in contact with the ground respectively. A time-based domain switching determines the number of feet in contact with the ground. When the QP fails in the DSP due to early lift-off or due to the delayed impact for the unplanned downstep scenario, a SSP controller is used as a backup controller.


%% file: 5_results.tex
\section{Results}
\label{sec:results}
\begin{figure*}[t]
    \centering
     \framebox{\includegraphics[width=0.98\textwidth]{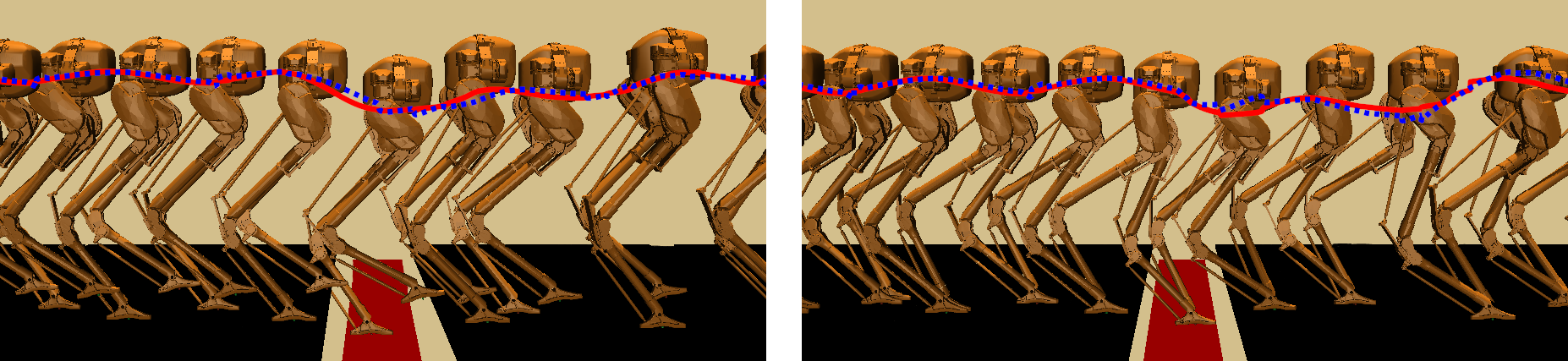}}
    \caption{Gait tiles (associated with the plots in Fig. \ref{fig:cassie_3D}) of Cassie walking down an planned downstep (left) and an unplanned downstep (right) of 10 centimeters. The downstep itself is indicated by the red tile. The desired and actual CoM trajectory are shown in red and blue respectively.}
    \label{fig:cassie_3D_tiles}
\end{figure*}


The methodology presented in this paper was applied to the aSLIP model and the robot Cassie to enable walking over planned and unplanned downsteps.

\subsection{aSLIP Walking}

We first apply the proposed approach to the aSLIP model.  
We successfully demonstrate the navigation of the downsteps for the RoM based on the stiffness and damping found from the optimization. The RoM representation of the human can therefore employ the kinematic and kinetic data from humans to overcome identical situations.
Although the vertical CoM behavior and the GRFs obtained form the human are successfully embedded onto the aSLIP walker, the fact that the aSLIP model assumes a massless swing leg prevents an accurate representation of the step-sizes of the human as shown in Fig. \ref{fig:human_BBF}. It is also shown that, due to the changes in vertical CoM position, the periodic orbits occur outside the orbital lines of the theoretical H-LIP model. For the planned case, this seems to increase the walking velocity during downstep while for the unplanned case, this leads to a decrease. This is in contrast to what we observe in the human data as shown in the supplementary video \cite{verhagen2022supplement}.

\subsection{3D Cassie Walking}
The main simulation results of the paper demonstrate the successful translation of human RoM data to the realization of downstep behaviors on Cassie (illustrated in Fig. \ref{fig:cassie_3D_tiles}).   
To achieve 3D walking, we are concerned with stabilizing the coronal plane which is successfully achieved with P2 orbits of the H-LIP \cite{xiong20213d}. The trade-off between tracking and force embedding becomes apparent in the downstepping scenarios.  In Fig. \ref{fig:cassie_3D} we see decreased tracking performance for the unplanned downstep scenario as the recovery behavior exceeds the one step required for the planned scenario. We argue that this is predominantly due to inaccurate desired GRFs from the human measurements. The control of the swing leg and vertical CoM for humans is mostly governed by the passive dynamics of the system when experiencing the unplanned downstep. For the robot, the vertical swing foot behavior is explicitly controlled at all times which prevents a fast impact velocity. With the requirement of lowering the CoM, this results in a minimum GRF during the downstep.

The improved controllability of Cassie w.r.t. the human, due to the leight-weight feet and instantaneous sensing of the downstep, means that for both planned and unplanned downsteps the increase of the forward CoM velocity is significantly reduced. Thus both planned and unplanned downsteps are traversed more effectively. The proposed controller based on embedding behavior based on the foot penetration help traverse unplanned changes in walking height and explicitly plan motion when exceeding nominal step-time.

%% file: 6_conclusions.tex
\section{Conclusion and Future Work}
We have shown that the human response to planned and unplanned downsteps can be captured by the aSLIP model which can similarly describe the dynamics of Cassie. Additionally, it was shown that the walking responses of the human to the environment can be embedded on morphologically and dynamically different robotic bipedal systems. By scaling the outputs of the human in the motion synthesis and embedding the contact forces in the low-level control, dynamic similarity between models is realized on the closed-loop systems. The proposed method has been successfully realized on an actuated SLIP model representative of the observed human and Cassie, as well as a 3D simulated Cassie to overcome planned and unplanned downsteps with similar responses to those found in the human gait. We have thus shown how human data can inspire the motion planning and the generation of reflex-like action for morphologically different bipedal walkers. 

The presented work currently focuses on a specific downstep scenario and realization in simulation. Future work will consider a general framework of transferring versatile human locomotion to dynamic bipedal robot behaviors and implementation on hardware. Additionally, our results from fitting the aSLIP model to the human are used in generating closed-loop stable walking on a RoM representative of the human, yet the resulting actuation signal itself is not used in this formulation. Future work could include realizing reflex-like action based on the results of this optimization. 

Lastly, the underlying issue which we preliminary address in this work is ensuring dynamic similarity between the RoM representations of the human and the robot. This could be further substantiated by specifically relating the representative leg stiffness and damping obtained from the optimization with the leg stiffness and damping of Cassie.

